\documentclass[runningheads]{llncs}

\usepackage[utf8]{inputenc}
\usepackage{hyperref}
\usepackage[misc]{ifsym}
\usepackage{graphicx}
\usepackage{wrapfig}
\usepackage{verbatim}
\usepackage{paralist,amsmath,color,enumerate,framed}
\usepackage{todonotes}

\title{An Ontology Design Pattern for Role-Dependent Names}
\author{Rushrukh Rayan\inst{1}\Letter \and Cogan Shimizu\inst{2} \and Pascal Hitzler\inst{1}}
\institute{Kansas State University, USA \and Wright State University, USA}
\authorrunning{Rayan, Rushrukh; Shimizu, Cogan; Hitzler, Pascal}
\begin{document}
\maketitle
\begin{abstract}
    We present an ontology design pattern for modeling \textsf{Names} as part of \textsf{Roles}, to capture scenarios where an \textsf{Agent} performs different \textsf{Roles} using different \textsf{Names} associated with the different Roles. Examples of an Agent performing a Role using different Names are rather ubiquitous, e.g., authors who write under different pseudonyms, or different legal names for citizens of more than one country. The proposed pattern is a modified merger of a standard Agent Role and a standard Name pattern stub. 
\end{abstract}
\section{Introduction}
\label{sec:intro}
We present an ontology design pattern for role-dependent names of agents that appears to be of rather ubiquitous importance but has not been described explicitly yet, to the best of our knowledge. The pattern is a relatively straightforward reification that modifies previously published patterns, as we discuss below. However we believe that even straightforward patterns such as this should be modeled carefully and provided to the public in the spirit of easy reuse and development of tool support, in the spirit of, e.g., the Modular Ontology Modeling (MOMo) methodology \cite{momo2023}, the Modular Ontology Design Library (MODL) \cite{modl}, and the COModIDE software framework \cite{ShimizuHH20}.

\begin{figure}[tb]
    \centering
    \includegraphics[width=.9\textwidth]{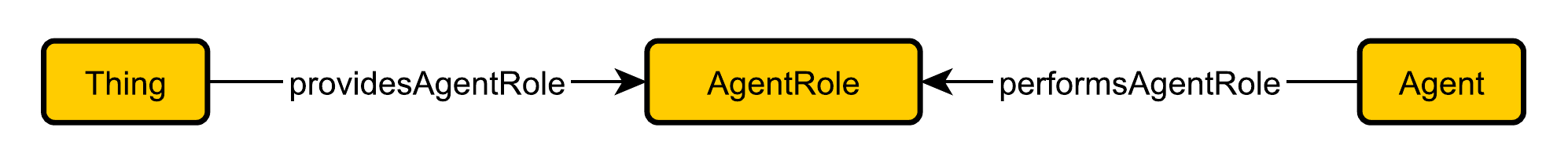}
    \includegraphics[width=.5\textwidth]{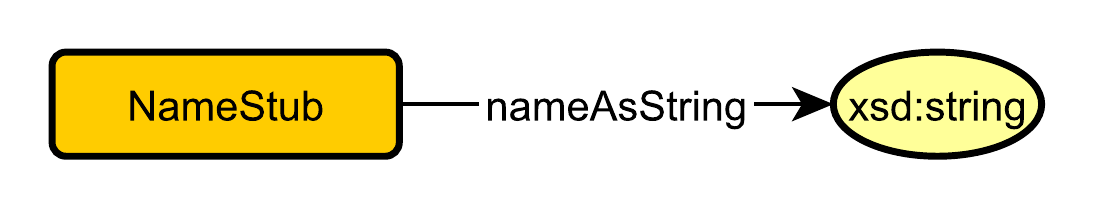}
    \caption{Diagrams for Agent Role pattern (top) and Name Stub pattern (bottom), as per \cite{modl}}
    \label{fig:agentrole}
\end{figure}

In the MODL library we find an AgentRole pattern as depicted in Figure \ref{fig:agentrole} (top), as well as a Name Stub pattern depicted in Figure \ref{fig:agentrole} (bottom).\footnote{Following MOMo \cite{momo2023}, we mostly discuss patterns by means of their schema diagrams, and only dive into axiomatization where needed, or at the very end when presenting the final pattern.} Typical uses of the former would be for roles of agents, such as being an employee in a company or author of a publication, where it is desirable to adorn the Role with additional context information like dates of employment or placement in the author sequence.

\begin{figure}[tb]
    \centering
    \includegraphics[width=\textwidth]{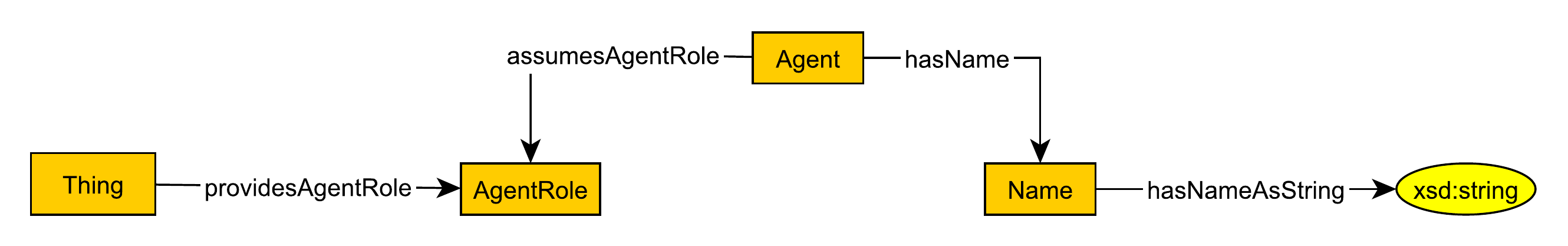}
    \caption{Diagram for naively joined AgentRole and NameStub patterns}
    \label{fig:agentrole-and-name}
\end{figure}

A schema diagram resulting from a na\"ive combination of these two patterns is depicted in Figure \ref{fig:agentrole-and-name} (with some mild renaming -- we are using patterns as templates as argued in \cite{HammarP16,momo2023}, rather than verbatim). Of course this pattern does not account for dependency of a name on the role: as an example for this type of dependency, consider the case of \emph{C. S. Lewis}, who published a collection of poems, \emph{Spirit in Bondage}, under the pseudonym \emph{Clive Hamilton}. On the other hand, his book entitled \emph{Grief Observed} was published under the pseudonym \emph{N. W. Clerk}. Using the diagram in Figure \ref{fig:agentrole-and-name} to naively encode this information would result in the triples in Figure \ref{fig:triples-naive} which, of course, do not convey which of the two (if in fact any) pseudonym was used for publication of which of the two books. Other example scenarios with essentially the same issue occur if persons or organizations have different legal names in different jurisdictions, may have their name changed at some stage, may use any type of occupational pseudonyms, or in the context of identity falsification using fake names.
\begin{figure}
\begin{verbatim}
:spiritInBondage :providesAgentRole :sibAuthorRole .
:griefObserved   :providesAgentRole :goAuthorRole .
:csLewis         :assumesAgentRole  :sibAuthorRole ,
                                    :goAuthorRole ;
                 :hasName           :csLewisNameCV ,
                                    :csLewisNameNWC ,
                                    :csLewisNameCSL .
:csLewisNameNWC  :hasNameAsString   "N. W. Clerk"^^xsd:string .
:csLewisNameCV   :hasNameAsString   "Clive Hamilton"^^xsd:string .
:csLewisNameCSL  :hasNameAsString   "C. S. Lewis"^^xsd:string .
\end{verbatim}
\caption{Example triples conforming to the diagram in Figure \ref{fig:agentrole-and-name}, for the example case of \emph{C. S. Lewis}, who published a collection of poems, \emph{Spirit in Bondage}, under the pseudonym \emph{Clive Hamilton}. On the other hand, his book entitled \emph{Grief Observed} was published under the pseudonym \emph{N. W. Clerk}}
\label{fig:triples-naive}
\end{figure}

The remainder of the paper is organized as follows. In Section~\ref{sec:pat} we present our pattern diagrammatically. In Section \ref{sec:axioms} we provide its axiomatic formalization. This is followed by conclusions in Section ~\ref{sec:conc}.

\section{Overview of the Role-Dependent Names Pattern}
\label{sec:pat}
\begin{figure}[h]
    \centering
    \includegraphics[width=\textwidth]{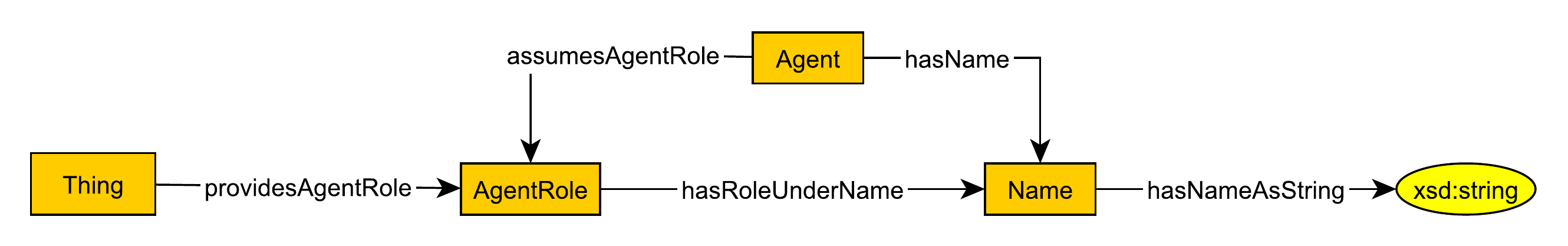}
    \caption{Schema Diagram for the Role-Dependent Names pattern}
    \label{fig:agentrolename}
\end{figure}

The difficulty posed by the diagram in Figure \ref{fig:agentrole-and-name} is, of course, easily addressed by making use of the fact that both AgentRole and Name are already reifications. The resulting diagram is depicted in Figure \ref{fig:agentrolename}. We will refer to this pattern as the Role-Dependent Names (in short, RDN) pattern.

Following this diagram, the example triples in Figure \ref{fig:triples-full} address the previously discussed \emph{C. S. Lewis} example. 

\begin{figure}[h]
\begin{verbatim}
:sibAuthorRole  :hasRoleUnderName  :csLewisNameCV .
:goAuthorRole   :hasRoleUnderName  :csLewisNameNWC .
\end{verbatim}
\caption{Additional triples, completing those in \ref{fig:triples-naive}, for the example case of \emph{C. S. Lewis}, who published a collection of poems, \emph{Spirit in Bondage}, under the pseudonym \emph{Clive Hamilton}. On the other hand, his book entitled \emph{Grief Observed} was published under the pseudonym \emph{N. W. Clerk}. The combined set of triples conforms with our Role-Dependent Name pattern, the schema diagram of which is depicted in Figure \ref{fig:agentrolename}.}
\label{fig:triples-full}
\end{figure}

\section{Pattern Axiomatization}
\label{sec:axioms}
Following the MOMo methodology \cite{momo2023}, we give a full set of axioms that we deem appropriate for the RDN pattern. The RDN pattern is driven by the interplay between the three core concepts: \textsf{Agent}, \textsf{AgentRole}, and \textsf{Name}. The OWLAx approach \cite{owlax,EberhartSCSH21}, which we follow here, suggests to first look at each node-edge-node ensemble in the schema diagram, and then at disjointness and additional axioms. Axiomatization is partially derived from the MODL library \cite{modl}.

\textsf{Agent} in the pattern generally refers to a person or an organization, i.e., inanimate objects would not usually fall under the scope of the Agent class. For instance, if a Table is used for dining, the Table cannot be thought of as an Agent that assumes the Role of a Dining Table in this scenario: it would still be a \emph{role}, but not an \emph{agent}role. 

With regard to axiomatization, it is natural to say that every agent must have a name (1). A structural tautology is used to convey that an Agent \emph{may} (but does not necessarily) also assume an AgentRole (2). To specify the domain of things that can assume an AgentRole, we make use of a Scoped Domain axiom to say that if something assumes an AgentRole, then it must be an Agent (3). 

Additionally, we use Inverse Functionality to restrict the number of agents that can assume a apecific AgentRole (4). Concretely, axiom (4) says that an AgentRole can be assumed by at most one Agent. Although this could be done differently, this choice constrains the shape of the RDF graph that complies with the pattern, and thus disambiguates the usage of the pattern.

\begin{align}
    \textsf{Agent} &\sqsubseteq \exists\textsf{hasName.Name} \\
    \textsf{Agent} &\sqsubseteq 
    \mathord{\geq}0\textsf{assumesAgentRole.AgentRole} \\
    \exists\textsf{assumesAgentRole.AgentRole} &\sqsubseteq \textsf{Agent}\\
    \textsf{AgentRole} &\sqsubseteq \leq \textsf{1assumesAgentRole$^{-}$.} \textsf{Agent}
\end{align} \medskip

\textsf{AgentRole}s mean the various roles an Agent can assume. We make use of a Scoped Range axiom to say that if an Agent assumes a Role, then it must be an AgentRole (5). 
Further, and centrally to this paper, an \textsf{AgentRole} may be assumed under a specific name, and we indicate this using a Structural Tautology axiom (6). The \textsf{hasRoleUnderName} property can furthermore be safely declared to have global range \textsf{Name}, and \textsf{providesAgentRole} can likewise be declared to have range \textsf{AgentRole}. 

\begin{align}
    \textsf{Agent} &\sqsubseteq \forall\textsf{assumesAgentRole.AgentRole} \\
    \textsf{AgentRole} &\sqsubseteq 
    \mathord{\geq}0\textsf{hasRoleUnderName.Name}\\
    \top &\sqsubseteq \forall\textsf{hasRoleUnderName.Name}\\
    \top &\sqsubseteq \forall\textsf{providesAgentRole.AgentRole}
\end{align} \medskip

For \textsf{Name}, Inverse Functionality is used to express that a Name can be the name of at most one Agent (9). This is done with a similar motivation as axiom~(4) above, i.e. to constrain the possible RDF graphs conforming with the pattern, i.e., to disambiguate use of the pattern. Axioms (10) and (11) declare global range for \textsf{hasName} and global domain for \textsf{hasNameAsString}, respectively.

\begin{align}
    \textsf{Name} &\sqsubseteq \leq \textsf{1hasName$^{-}$.} \textsf{Agent}\\
    \top &\sqsubseteq \forall\textsf{hasName.Name} \\
    \exists\textsf{hasNameAsString.} \top &\sqsubseteq \textsf{Name} 
\end{align} \medskip

We add the obvious disjointness axioms (12--14), and then also two role chains axioms, (15) and (16). (15) formalizes that if an agent assumes a role under a Name, then the Agent must have the \textsf{same} name. Similarly, (16) formalizes that if an agent has a name and a role is assumed under that name, then the agent must assume the same Role.

\begin{align}
    \textsf{AgentRole} \sqcap \textsf{Agent} &\sqsubseteq \bot \\
    \textsf{Agent} \sqcap \textsf{Name} &\sqsubseteq \bot \\
    \textsf{Name} \sqcap \textsf{AgentRole} &\sqsubseteq \bot \\
        \textsf{assumesAgentRole} \circ \textsf{hasRoleUnderName} &\sqsubseteq \textsf{hasName} \\
    \textsf{hasName} \circ \textsf{hasRoleUnderName}^- &\sqsubseteq \textsf{assumesAgentRole} 
\end{align}

\section{Conclusion}
\label{sec:conc}
The Role-Dependent Name pattern is aimed towards situations where there is an association between the Name and the Role as Agents can assume different roles under different names. As usual, the pattern is not meant to be rigid, in the sense that part of the pattern can be omitted when making use of it, or it can be extended as needed. E.g., AgentRoles may carry additional information such as spatio-temporal extents, and names may have a rich structure. 

\medskip

\noindent\emph{Acknowledgement.} The authors acknowledge funding under the National Science Foundation grants 2119753 "RII Track-2 FEC: BioWRAP (Bioplastics With Regenerative Agricultural Properties): Spray-on bioplastics with growth synchronous decomposition and water, nutrient, and agrochemical management" and 2033521: "A1: KnowWhereGraph: Enriching and Linking Cross-Domain Knowledge Graphs using Spatially-Explicit AI Technologies."
\bibliographystyle{splncs04}
\bibliography{refs}
\end{document}